\documentclass[11pt, a4paper, logo, copyright, nonumbering]{paddleocr}
\usepackage[authoryear, sort&compress, round]{natbib}
\usepackage{dblfloatfix}
\usepackage{ulem}
\usepackage{caption}
\usepackage{dramatist}
\usepackage{xspace}
\usepackage{pifont} %
\usepackage{multirow}
\usepackage{tcolorbox}
\usepackage{xltabular}
\usepackage{longtable}
\usepackage[hang, flushmargin]{footmisc}

\hypersetup{
    colorlinks=true,
    linkcolor=blue,
    citecolor=blue,
    filecolor=blue,
    urlcolor=blue,
    pdfpagemode=UseNone,
    pdfstartview=FitH
}
\usepackage{threeparttable} 
\usepackage{amsfonts}
\usepackage{amsmath}
\usepackage{amssymb}
\usepackage{lineno}
\usepackage{multirow}
\usepackage{adjustbox}

\usepackage{CJKutf8}
\usepackage{subfigure}
\usepackage{setspace}

\usepackage{dsfont}
\usepackage{array} %
\usepackage{tabularx} %
\usepackage{subfigure} %
\usepackage{xcolor} %
\usepackage{tabularx}
\usepackage{booktabs}

\usepackage{lipsum}  %
\usepackage{multicol} %

\usepackage{listings}
\usepackage{xcolor}

\usepackage{tablefootnote}
\usepackage{hyperref}

\lstdefinelanguage{json}{
    basicstyle=\ttfamily\footnotesize,
    numbers=left,
    numberstyle=\tiny\color{gray},
    stepnumber=1,
    numbersep=8pt,
    showstringspaces=false,
    breaklines=true,
    frame=lines,
    backgroundcolor=\color{gray!10},
    morestring=[b]",
    literate=
     *{0}{{{\color{black}0}}}{1}
      {1}{{{\color{black}1}}}{1}
      {2}{{{\color{black}2}}}{1}
      {3}{{{\color{black}3}}}{1}
      {4}{{{\color{black}4}}}{1}
      {5}{{{\color{black}5}}}{1}
      {6}{{{\color{black}6}}}{1}
      {7}{{{\color{black}7}}}{1}
      {8}{{{\color{black}8}}}{1}
      {9}{{{\color{black}9}}}{1}
}

\makeatletter
\def\@BTrule[#1]{%
  \ifx\longtable\undefined
    \let\@BTswitch\@BTnormal
  \else\ifx\hline\LT@hline
    \nobreak
    \let\@BTswitch\@BLTrule
  \else
     \let\@BTswitch\@BTnormal
  \fi\fi
  \global\@thisrulewidth=#1\relax
  \ifnum\@thisruleclass=\tw@\vskip\@aboverulesep\else
  \ifnum\@lastruleclass=\z@\vskip\@aboverulesep\else
  \ifnum\@lastruleclass=\@ne\vskip\doublerulesep\fi\fi\fi
  \@BTswitch}
\makeatother

\addto\extrasenglish{
}

 {\begin{list}{}%
         {\setlength{\leftmargin}{#1}}%
         \item[]%
 }
 {\end{list}}
 
\reportnumber{001} %

\renewcommand{\today}{}

\title{\centering PaddleOCR~3.0 Technical Report}

\author[*]{
\small
Cheng Cui, Ting Sun, Manhui Lin, Tingquan Gao, Yubo Zhang, Jiaxuan Liu,
\vspace{-0.4cm}
\\
\small
 Xueqing Wang, Zelun Zhang, Changda Zhou, Hongen Liu, Yue Zhang, Wenyu Lv, 
\\
\small
Kui Huang, Yichao Zhang, Jing Zhang, Jun Zhang, Yi Liu, Dianhai Yu, Yanjun Ma
\vspace{0.2cm}
\\
\small
\textbf{PaddlePaddle Team, Baidu Inc.}
\\
\small
\texttt{paddleocr@baidu.com}
\vspace{0.2cm}
  \\
  {\small
  \raggedright{  % 激活左对齐
  \small
  \hspace{7.06em}  % 重置默认缩进
  \includegraphics[height=0.9em]{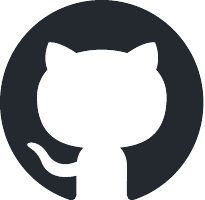} \textbf{Source Code}: \url{https://github.com/PaddlePaddle/PaddleOCR} \\
  \hspace{-1.2em}  % 重置默认缩进
  \small
  \includegraphics[height=0.9em]{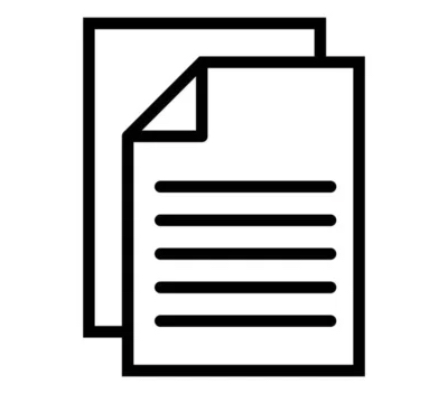} \textbf{Document}: \url{https://paddlepaddle.github.io/PaddleOCR} \\
  \small
  \hspace{1.9em}  % 重置默认缩进
  \includegraphics[height=1.0em]{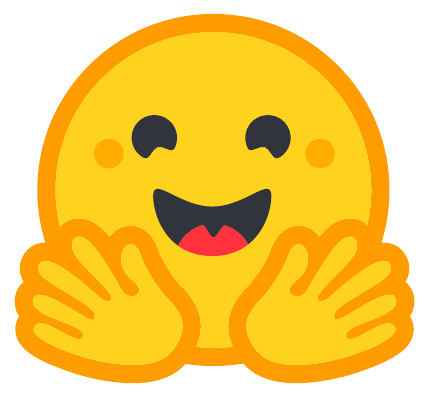} \textbf{Models \& Online Demo}: \url{https://huggingface.co/PaddlePaddle} \\
  \small
  }

  }
}

\renewcommand{\phi}{\varphi}

\renewcommand{\epsilon}{\varepsilon}
\renewcommand{\imath}{\mathrm{i}}

\newlength{\restsubwidth}
\newlength{\restsubheight}
\newlength{\restsubmoreheight}
\newcommand{\rest}[2]{%
        \settowidth{\restsubwidth}{\ensuremath{#2}}
        \settoheight{\restsubheight}{\ensuremath{{}_{#2}}}
        \ensuremath{{#1\hskip 0.5pt}_{\vrule\kern2pt\parbox[b][%
        4pt][b]{\the\restsubwidth}{%
                        \ensuremath{{}_{#2}}}}}
        }

\begin{abstract}

This technical report introduces PaddleOCR~3.0, an Apache-licensed open-source toolkit for OCR and document parsing. To address the growing demand for document understanding in the era of large language models, PaddleOCR~3.0 presents three major solutions: (1) PP-OCRv5 for multilingual text recognition, (2) PP-StructureV3 for hierarchical document parsing, and (3) PP-ChatOCRv4 for key information extraction. Compared to mainstream vision-language models (VLMs), these models with fewer than 100 million parameters achieve competitive accuracy and efficiency, rivaling billion-parameter VLMs. In addition to offering a high-quality OCR model library, PaddleOCR~3.0 provides efficient tools for training, inference, and deployment, supports heterogeneous hardware acceleration, and enables developers to easily build intelligent document applications.

\end{abstract}

\begin{document}

\maketitle

\begin{figure}[h]
    \centering
    \subfigure[PP-OCRv5]{\includegraphics[height=.18\textheight]{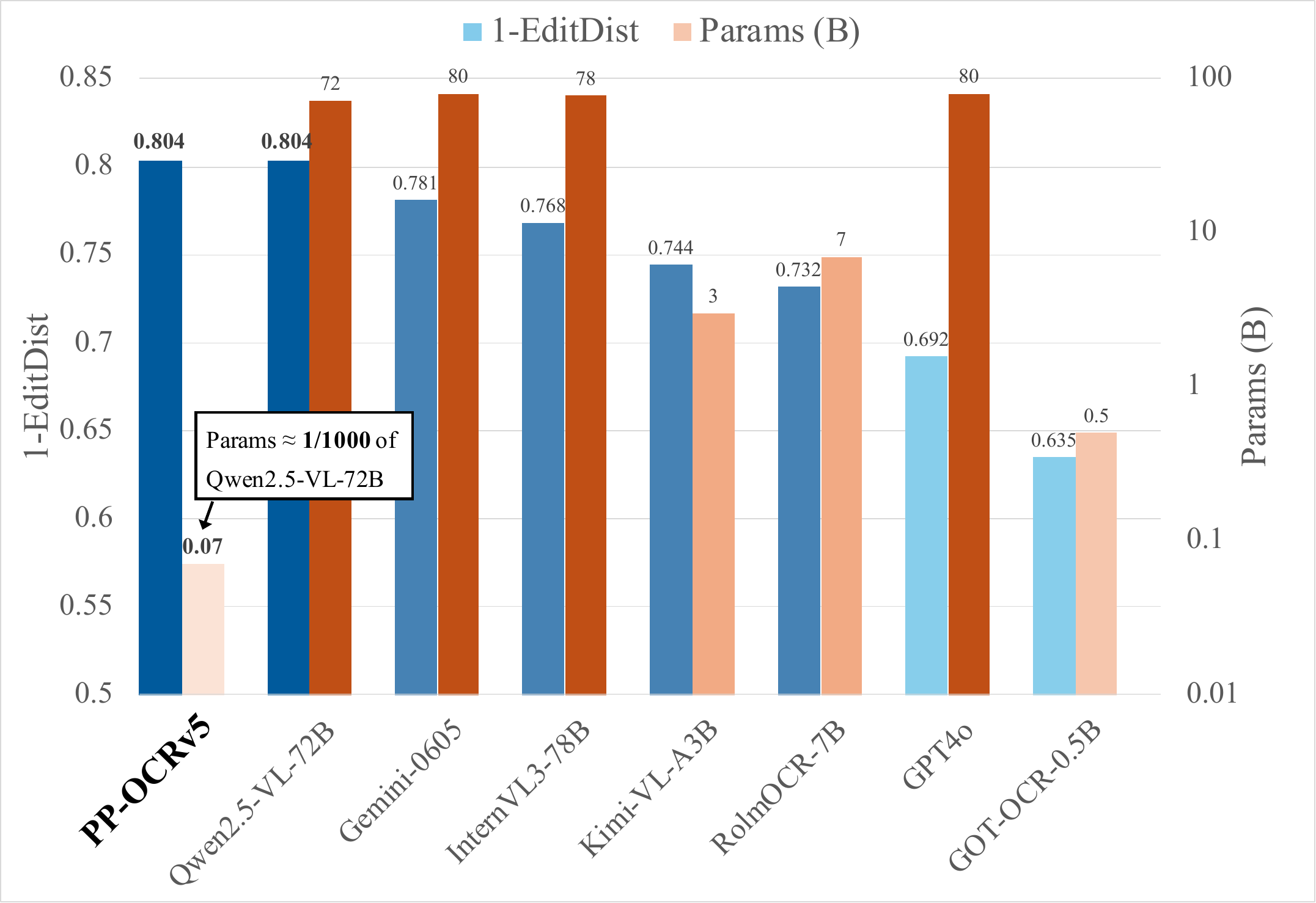}}
    \subfigure[PP-StructureV3]{\includegraphics[height=.18\textheight]{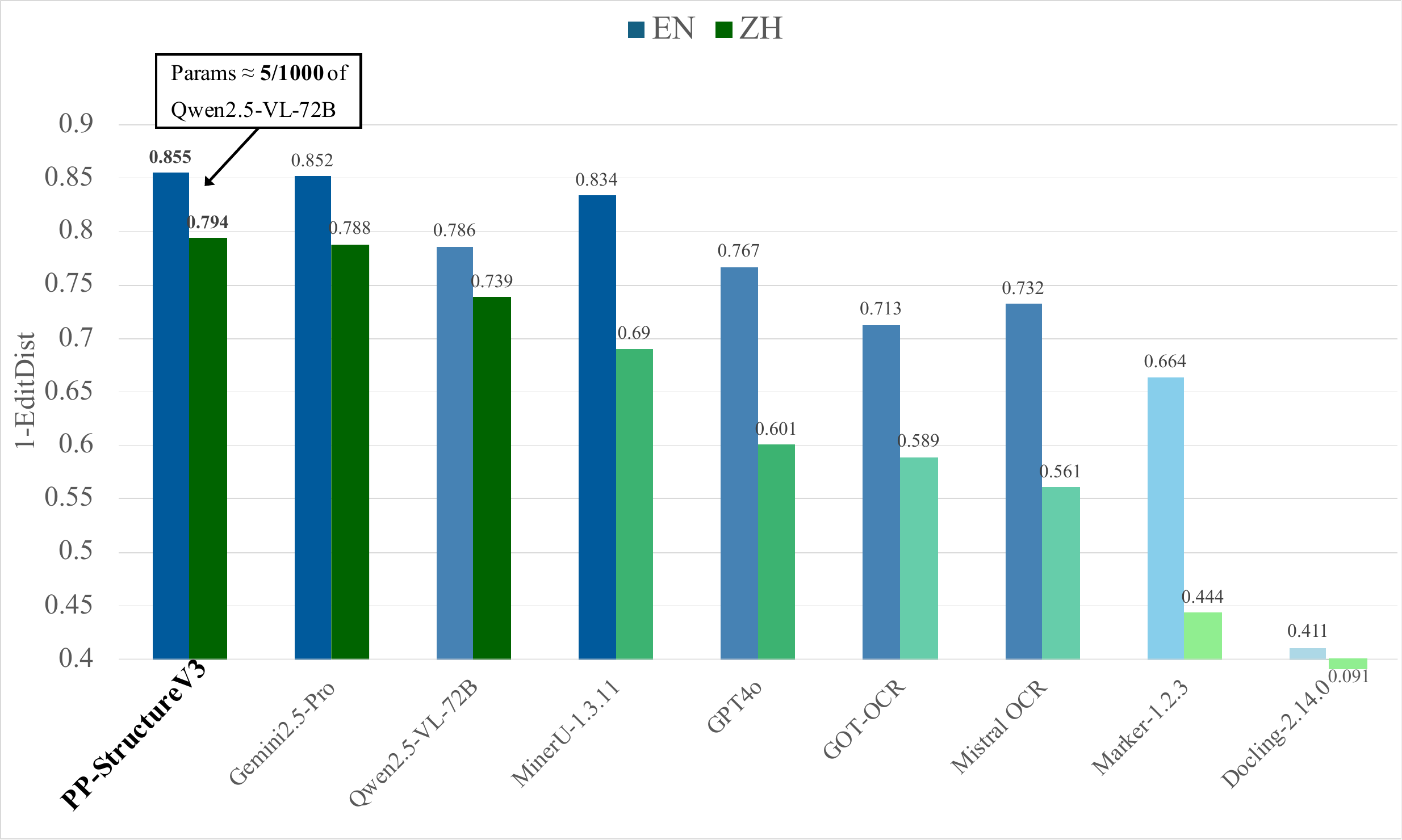}}
    \caption{Performance comparison of PP-OCRv5 and PP-StructureV3 with their respective counterparts. The evaluation set for PP-OCRv5 is our self-built dataset, which includes multiple writing formats such as Simplified Chinese, Traditional Chinese, Chinese Pinyin, English, and Japanese. The evaluation set for PP-StructureV3 is OmniDocBench~\citep{ouyang2025omnidocbench}. The term "1-EditDist" refers to \(1 - \text{Edit Distance}\), where a higher value indicates better performance.
}
    \label{fig:benchmark_performance}
\end{figure}

% \newpage

% \begin{spacing}{0.9}
% \tableofcontents
% \end{spacing}

\newpage

\section{Introduction}

Optical Character Recognition (OCR) is a foundational technology that enables the conversion of images containing text, scanned documents into structured, machine-readable text. Its significance has never been more pronounced than in the current era of artificial intelligence, where massive volumes of unstructured visual data are generated and consumed daily across scientific, industrial, and social domains. The recent surge in large language models (LLMs)\citep{achiam2023gpt,guo2025deepseek,yang2025qwen3,ernie2025technicalreport} and Retrieval-Augmented Generation (RAG) systems\citep{lewis2020retrieval} has further elevated the strategic importance of OCR: it is no longer sufficient for OCR systems to merely transcribe text accurately---they must now serve as critical enablers in the construction of high-quality datasets, facilitate knowledge extraction, and act as bridges between the visual and semantic layers of modern AI systems.

The evolution of OCR technology reflects the broader trajectory of computer vision and natural language processing. Early OCR systems\citep{mori1999ocr,506792}, based on hand-crafted features and rule-based heuristics, performed adequately under controlled conditions but quickly reached their limits when confronted with the complexity and diversity of real-world scenarios. The advent of deep learning, particularly convolutional neural networks (CNNs) and their derivatives, ushered in a new era of data-driven OCR, enabling substantial improvements in recognition accuracy, robustness, and adaptability\citep{goodfellow2014multidigitnumberrecognitionstreet,shi2015endtoendtrainableneuralnetwork}. However, as large-scale AI applications proliferate, new requirements have emerged: OCR engines need to handle a broader range of documents---from handwritten notes and multilingual content to rare or historical scripts, and complex layouts with tables, charts, and embedded images. Furthermore, in industrial and research settings alike, OCR is increasingly expected to support downstream tasks such as document understanding, key information extraction (KIE), and semantic search, often as part of end-to-end intelligent workflows.

In recent years, the rapid advancement of LLMs and RAG systems has fundamentally transformed the landscape of information retrieval and knowledge management. These systems rely heavily on the availability of high-quality, diverse, and accurately labeled textual corpora for both pre-training and inference. OCR, in this context, is not simply a data acquisition tool, but a linchpin technology that fuels the entire pipeline---from digitizing vast archives of scientific literature to enabling real-time question answering over heterogeneous document collections. The accuracy and comprehensiveness of OCR outputs directly influence the performance and trustworthiness of LLM-based applications, especially in domains where information is predominantly shared in scanned or image-based formats (e.g., legal documents, historical records, scientific papers, and business forms). Moreover, RAG architectures, which combine retrieval mechanisms with generative modeling, are particularly sensitive to the quality of underlying document representations. Inadequate OCR can propagate errors, introduce noise, or omit critical content, thereby undermining the effectiveness of retrieval and the factual correctness of generated responses.

Despite these pressing needs, existing OCR solutions still face significant challenges in practical deployment. Traditional pipelines typically struggle with low-quality scans, complex backgrounds, non-standard fonts, and multi-modal documents that blend text with figures, tables, or handwritten annotations. The diversity of real-world languages, scripts, and writing styles further complicates the recognition process, requiring not only robust visual modeling but also powerful language understanding capabilities. In addition, industrial and research users increasingly demand lightweight, easily deployable solutions that can be adapted to different hardware constraints and integrated seamlessly with larger AI ecosystems. The open-source community has played a pivotal role in democratizing access to advanced OCR technology, yet there remains a gap between academic research prototypes and production-ready systems capable of supporting the stringent requirements of dataset construction, RAG workflows, and large-scale document intelligence.

\textbf{PaddleOCR 1.x \& 2.x: Advancements and Innovations in Open-Source OCR Technology}

PaddleOCR has emerged as a prominent open-source project addressing these multifaceted challenges. Since its initial release in 2020, PaddleOCR has adhered to the principles of comprehensive coverage, end-to-end workflow, and lightweight efficiency, setting new standards for both usability and technical excellence in the OCR domain. Anchored by the PP-OCR series, PaddleOCR has evolved through multiple iterations---each pushing the boundaries of text detection, recognition, and document analysis. Early versions such as PP-OCRv1\citep{du2020pp} focused on achieving an optimal balance between accuracy and speed, making OCR accessible for resource-constrained environments. Subsequent releases (PP-OCRv2\citep{du2021pp}, v3\citep{li2022pp}, and v4) incrementally improved recognition performance, extended language coverage, and introduced sophisticated models for handwriting and rare character recognition. A notable advancement has been the integration of document structural understanding via the PP-Structure series, enabling PaddleOCR to move beyond text lines and paragraphs to address complex layout analysis, table structure recognition (e.g., SLANet\citep{li2022ppstructurev2}), and other advanced parsing tasks. These capabilities have made PaddleOCR a critical engine for automated document processing, intelligent archiving, information extraction, and, increasingly, for supporting the data pipelines of LLMs and RAG systems.

The adoption and impact of PaddleOCR in both academic and industrial communities are evidenced by its widespread use and vibrant developer ecosystem. With more than 50,000 stars on GitHub as of June 2025, and its deployment as the core OCR engine in projects such as MinerU\citep{wang2024mineru}, RAGFlow\citep{ragflow}, and UmiOCR\citep{UmiOCR}, PaddleOCR has become an indispensable tool for digitization initiatives, knowledge management platforms, and AI-driven document analysis workflows. Notably, PaddleOCR has played a central role in the construction of high-quality document datasets for large model training, enabling researchers to assemble diverse, accurately annotated corpora spanning multiple languages, domains, and document types. Its modular architecture and rich API ecosystem facilitate seamless integration with RAG pipelines, where efficient and accurate OCR is essential for document ingestion, retrieval indexing, and context provision to generative models.

As PaddleOCR’s user base has expanded, so has the range of feedback and requirements from the community. Users have highlighted persistent needs in areas such as robust handwriting recognition, improved support for multi-language and rare script recognition, more powerful document parsing for complex layouts, and advanced key information extraction. These demands are further amplified by the growing scale and dynamism of LLM and RAG applications, where the ability to extract, structure, and semantically interpret information from diverse documents is a prerequisite for building reliable, responsive, and intelligent systems. Aware of these trends and our responsibility as a leading open-source platform, we remain committed to continuously improving PaddleOCR to meet the evolving challenges of the field.

\begin{figure}[h]
\centering
\includegraphics[width=1.0\textwidth]{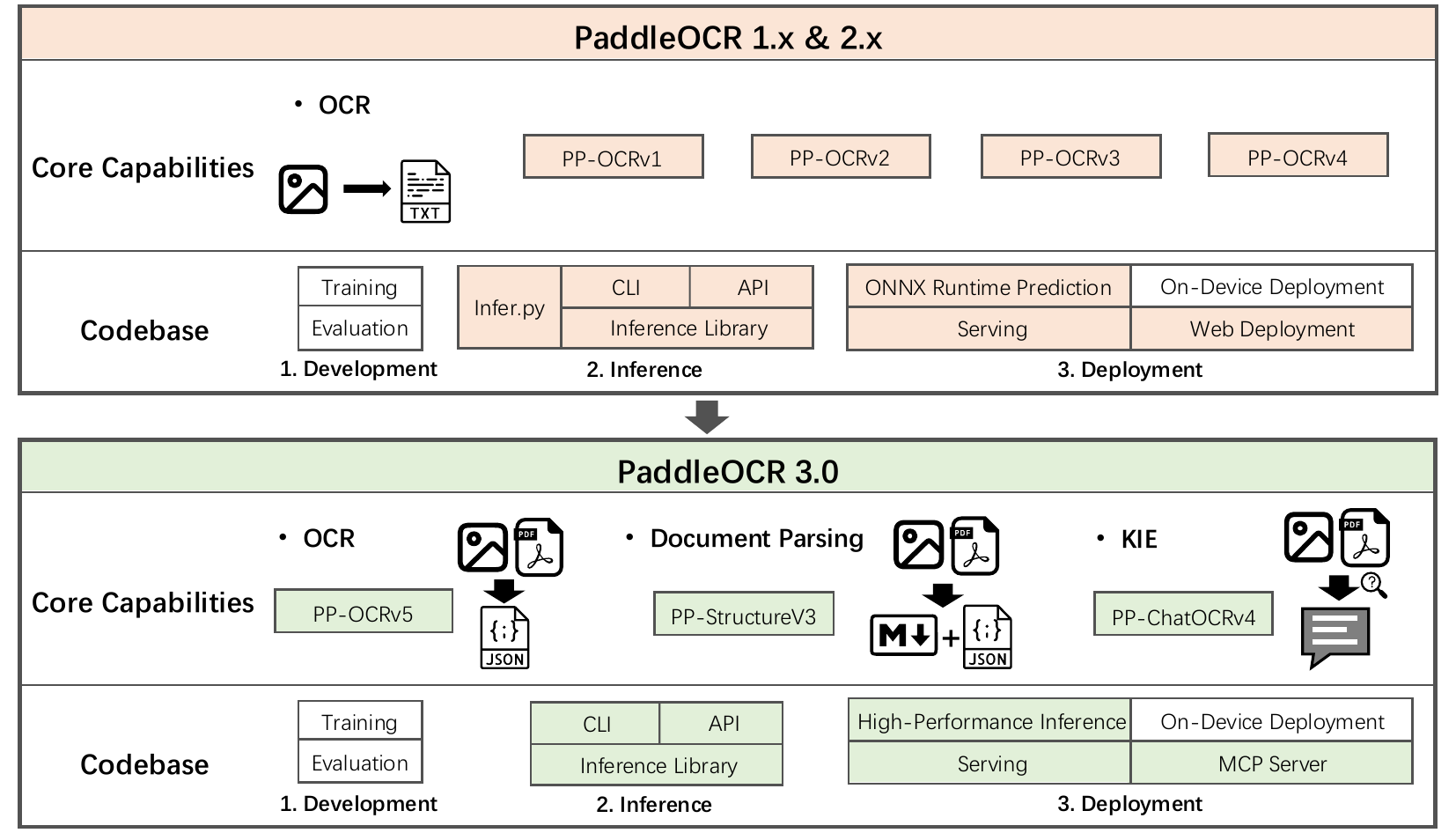}
\caption{
    \centering
     Evolution from PaddleOCR 1.x \& 2.x to PaddleOCR 3.0. Different colors have been employed to denote areas with notable discrepancies between PaddleOCR 1.x \& 2.x and PaddleOCR 3.0.
}
\label{fig:paddleocr_upgrade}
\end{figure}

\textbf{PaddleOCR 3.0: A New Milestone in Enhancing Text Recognition and Document Parsing}

In this context, we introduce PaddleOCR~3.0, a major release designed to systematically enhance text recognition accuracy and document parsing capabilities, with a particular focus on the complex scenarios encountered in modern AI applications. PaddleOCR~3.0 encompasses several core innovations. First, it presents the high-precision text recognition pipeline PP-OCRv5, which leverages advanced model architectures and training strategies to deliver state-of-the-art accuracy across printed, handwritten, and multilingual documents, while maintaining efficiency suitable for both cloud and edge deployment. Moreover, PP-OCRv5 achieves unified recognition of Simplified Chinese, Traditional Chinese, Chinese Pinyin, English, and Japanese within a single model. Second, PaddleOCR~3.0 includes PP-StructureV3, a document parsing solution that integrates layout analysis, table recognition, and structure extraction in an end-to-end framework, enabling accurate and scalable document understanding for forms, invoices, scientific literature, and more. Third, recognizing the need for deeper semantic integration, we introduce PP-ChatOCRv4, a system that combines lightweight OCR models with large language models to facilitate key information extraction, context-aware question answering, and flexible document comprehension—capabilities that are essential for powering RAG pipelines and intelligent document agents. In addition, PaddleOCR~3.0 extends its coverage with dedicated solutions for specialized tasks such as seal text recognition, formula recognition, and chart analysis, further expanding its utility for both research and industrial use cases.

The contributions of PaddleOCR~3.0 are not limited to technical innovation; the release continues to prioritize openness, usability, and extensibility, with a robust API ecosystem, comprehensive model zoo, and active community support. In this version, several legacy design flaws have been revised to provide cleaner and extensible API and CLI, while maintaining a reasonable degree of backward compatibility. At the deployment level, PaddleOCR~3.0 has been restructured to deliver a more streamlined, out-of-the-box experience and improved integration capabilities with LLMs. By targeting key challenges in complex OCR scenarios and aligning with the fundamental needs of data construction for LLMs and RAG pipelines, PaddleOCR~3.0 aspires to serve as an efficient, intelligent, and open infrastructure for document AI. We hope that this advancement will accelerate the development of intelligent automation and knowledge-driven AI systems, foster new research and application frontiers at the intersection of vision and language, and promote document processing toward higher levels of intelligence and automation.

\section{Core Capabilities}

PaddleOCR 3.0 comprises three core capabilities: PP-OCRv5, PP-StructureV3 and PP-ChatOCRv4. This section elaborates on the problems addressed by these capabilities, details of the model solution, and their performance metrics.

\subsection{PP-OCRv5}\label{sec:pp-ocrv5} 

\begin{figure}[h]
\centering
\includegraphics[width=1.0\textwidth]{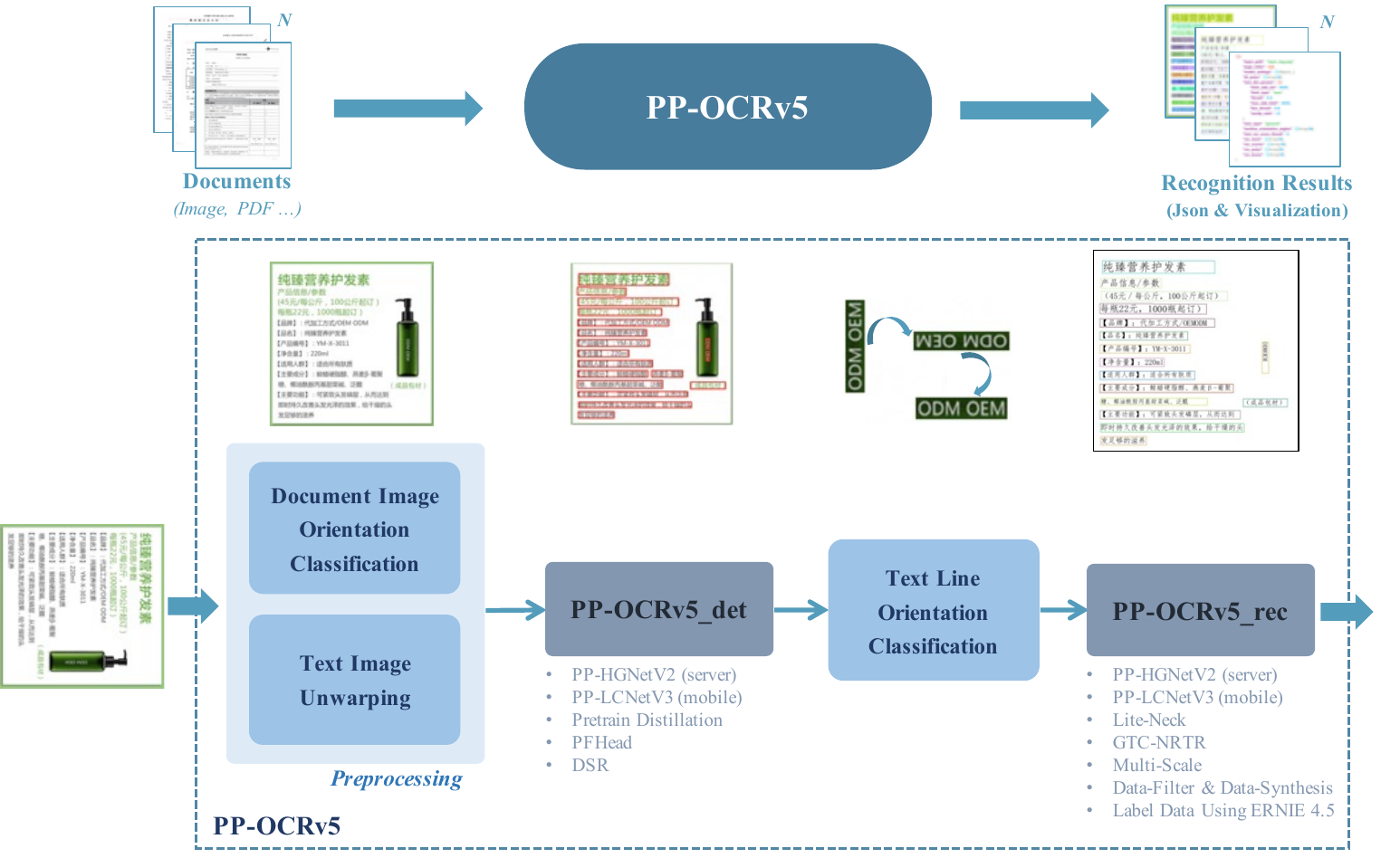}
\caption{
    \centering
    Pipeline of PP-OCRv5. The pipeline includes image preprocessing, text region detection, text line orientation classification, and text recognition, ultimately extracting the text from images and outputting it as structured textual content. 
}
\label{fig:ppocrv5_pipeline}
\end{figure}

PP-OCRv5 is a high-precision and lightweight OCR system designed to perform effectively in a wide range of scenarios. It supports a diverse range of scripts within a single model, including Simplified Chinese, Traditional Chinese, Chinese Pinyin, English, and Japanese. To address the diverse hardware environments and varying requirements for inference speed, PP-OCRv5 offers two distinct model variants: a server version and a mobile version. The server version is specifically optimized for systems equipped with hardware accelerators such as GPUs, thereby enabling accelerated inference and higher throughput. In contrast, the mobile version is tailored for deployment in CPU-only environments, with optimizations targeting resource-constrained devices. Unless otherwise specified, all mentions of PP-OCRv5 in this paper refer by default to the server version. Figure~\ref{fig:ppocrv5_pipeline} illustrates the framework of PP-OCRv5, which comprises four key components: image preprocessing, text detection, text line orientation classification, and text recognition. The following will introduce these four components.

1. \textbf{Image Preprocessing Module}: The image preprocessing module is crucial for preparing the input images by enhancing their quality and adjusting distortions or orientation issues. This process lays a solid foundation for accurate text detection and recognition. PP-OCRv5 includes an optional image preprocessing module to handle image rotation and geometric distortion. This module includes an image orientation classification model based on PP-LCNet~\citep{cui2021pplcnetlightweightcpuconvolutional} and a text image unwarping model built on UVDoc~\citep{verhoeven2023uvdoc}. Users can choose to use these features based on their application scenarios.

2. \textbf{Text Detection Model}: The PP-OCRv5 text detection model enhances its predecessor, PP-OCRv4, through optimizations in three key aspects: network architecture, distillation strategy, and data augmentation. Firstly, PP-OCRv5 adopts the more advanced PP-HGNetV2\footnote{\label{hgnetv2}\url{https://github.com/PaddlePaddle/PaddleClas/blob/release/2.6/docs/en/models/PP-HGNetV2_en.md}} as its backbone network, replacing the previous PP-HGNet.  Moreover, PP-OCRv5 enhances model robustness through knowledge distillation, utilizing a visual encoder from the advanced GOT-OCR2.0~\citep{wei2024general} as the teacher model to transfer knowledge by aligning feature representations with the PP-HGNetV2 student. Finally, PP-OCRv5 detection model enhances text detection performance by incorporating advanced data augmentation techniques, including hard case mining via ERNIE-4.5-VL-424B-A47B comparison and text line-based multilingual strategies (random synthesis, rotation, blurring, geometric transformations). Notably, PP-OCRv5 detection model retains effective strategies from PP-OCRv4, such as the PFHead (Parallel Fusion Head) architecture and the DSR (Dynamic Scale-aware Refinement) training strategy. Overall, PP-OCRv5 improves the model’s ability to generalize across diverse datasets, leading to more accurate and reliable text detection in real-world scenarios. 

%This upgraded backbone achieves a superior balance between performance and efficiency, delivering significantly faster inference speeds across various precision levels.

3. \textbf{Text Line Orientation Classification Model}: In PP-OCRv5, the text line orientation classification model is primarily responsible for determining and correcting the orientation of detected text lines. When the input text lines are misoriented (e.g., inverted or rotated), this model can automatically identify and rectify their direction to ensure that they are in the standard readable orientation. This step guarantees that the subsequent text recognition model receives a correctly oriented text line, thereby improving the overall accuracy and robustness of the OCR system.

4. \textbf{Text Recognition Model}: The PP-OCRv5 recognition model employs a dual-branch architecture with PP-HGNetV2 as the backbone: one branch (GTC-NRTR) uses attention-based training to enhance sequence modeling~\citep{hu2020gtc}, while the other (SVTR-HGNet) focuses on efficient inference with CTC loss. During training, the GTC-NRTR branch guides the SVTR-HGNet branch~\citep{du2022svtr}, but only the lightweight SVTR-HGNet branch is used for prediction, ensuring both accuracy and speed~\citep{li2022ppocrv3}. For data construction, PP-OCRv5 combines traditional models with the ERNIE-4.5-VL-424B-A47B to automatically annotate and filter high-quality handwritten samples, including rare characters generated through synthesis. Additionally, large-scale labeled data is obtained from documents like PDFs and e-books using automated parsing and edit distance filtering. These data construction strategies also provide a solid data foundation for the overall performance improvement of PP-OCRv5.

\begin{figure}[h]
\centering
\includegraphics[width=1.0\textwidth]{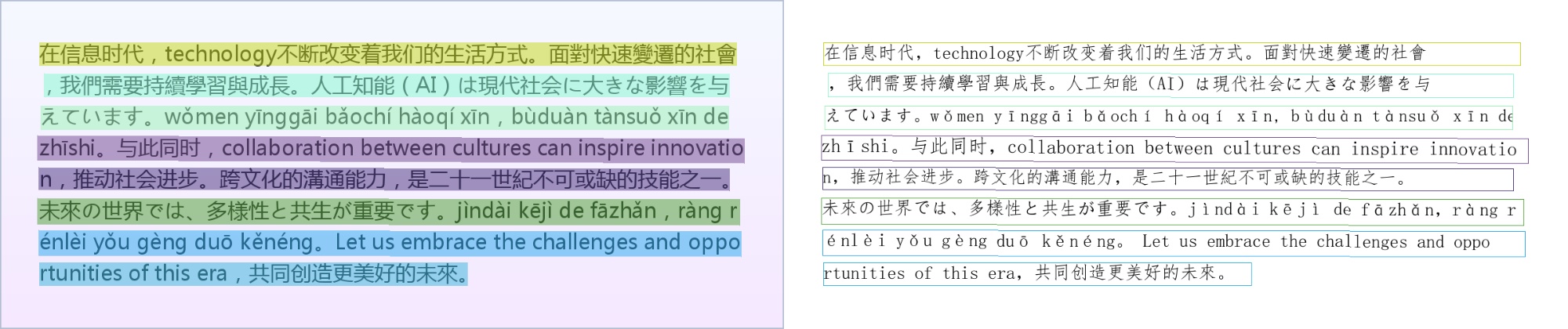}
\caption{
    \centering
    Multilingual recognition example.
}
\label{fig:ocrv5example1_1}
\end{figure}

Accordingly, the core contributions of PP-OCRv5 are as follows:

1. \textbf{Unified Multilingual Modeling}: PP-OCRv5 achieves unified recognition of Simplified Chinese, Traditional Chinese, Chinese Pinyin, English, and Japanese within a single model. Through an innovative unified architecture design, the model maintains a compact size under 100~MB. This resolves efficiency bottlenecks caused by integrating multiple models in multilingual scenarios, significantly simplifying industrial deployment processes. An example illustrating PP-OCRv5's multilingual recognition performance is shown in Figure~\ref{fig:ocrv5example1_1}.

\begin{figure}[h]
\centering
\includegraphics[width=1.0\textwidth]{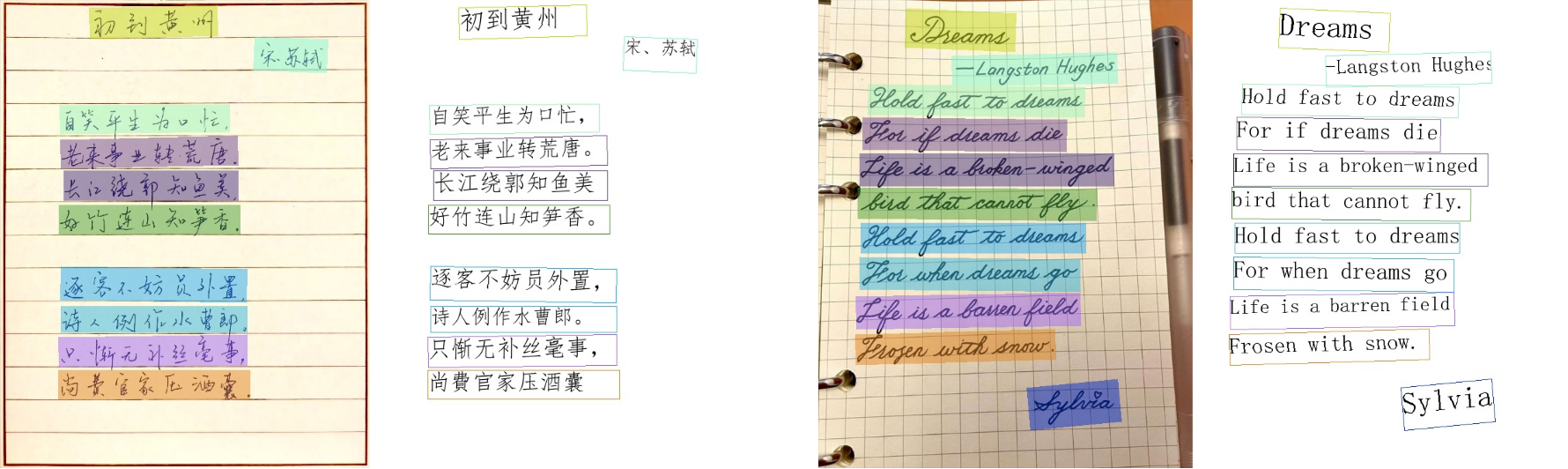}
\caption{
    \centering
    Handwritten Chinese characters (left) and handwritten English text (right). 
}
\label{fig:ocrv5example2}
\end{figure}

2. \textbf{Robust Recognition of Complex Handwriting}: To address the demands of key application domains such as examination grading in education, bill recognition in finance, and contract entry in the legal sector, PP-OCRv5 has significantly enhanced its capability in handwritten text recognition. Experimental results demonstrate that, compared to previous models, PP-OCRv5 reduces the recognition error rate by 26\% on tasks involving non-standard handwriting forms, including both Chinese and English handwritten texts. Figure~\ref{fig:ocrv5example2} illustrates the performance of PP-OCRv5 in handwritten text recognition.

%Targeting the critical needs in education (exam grading), healthcare (medical record digitization), and legal (contract transcription) sectors, PP-OCRv5 introduces an adaptive feature decoupling mechanism. This approach reduces recognition error rates by 37\% compared to previous models in non-standard writing forms such as Chinese cursive, English script, and Japanese variants. Figure~\ref{fig:ocrv5example2} demonstrates PP-OCRv5’s handwritten text recognition capabilities.

\begin{figure}[h]
\centering
\includegraphics[width=1.0\textwidth]{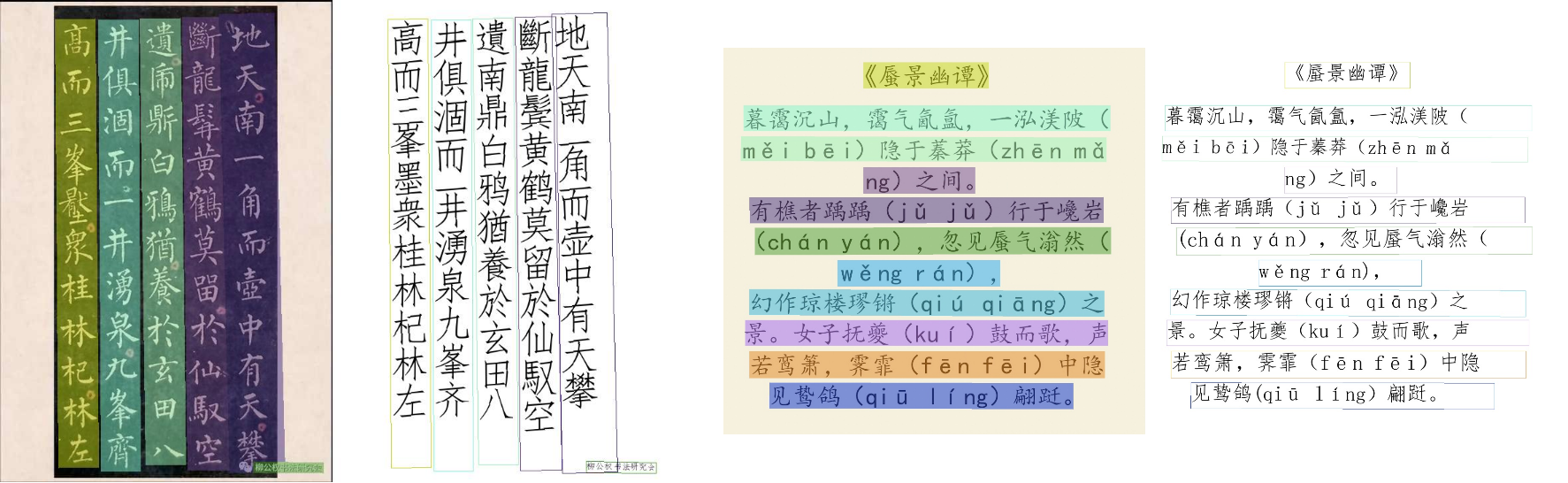}
\caption{
    \centering
    Vertical Chinese ancient book text (left) and rare Chinese character (right).
}
\label{fig:ocrv5example3_2}
\end{figure}

3. \textbf{Robust Recognition of Historical Texts and Uncommon Characters}: In complex and non-standard writing scenarios such as Chinese ancient texts and rare Chinese characters, PP-OCRv5 significantly improves text recognition accuracy through optimization of network architecture and systematic construction of diverse, high-quality datasets, effectively meeting the high-precision text recognition requirements across various domains. The recognition performance in complex scenarios is shown in Figure~\ref{fig:ocrv5example3_2}.

\begin{figure}[h]
\centering
\includegraphics[width=0.99\textwidth]{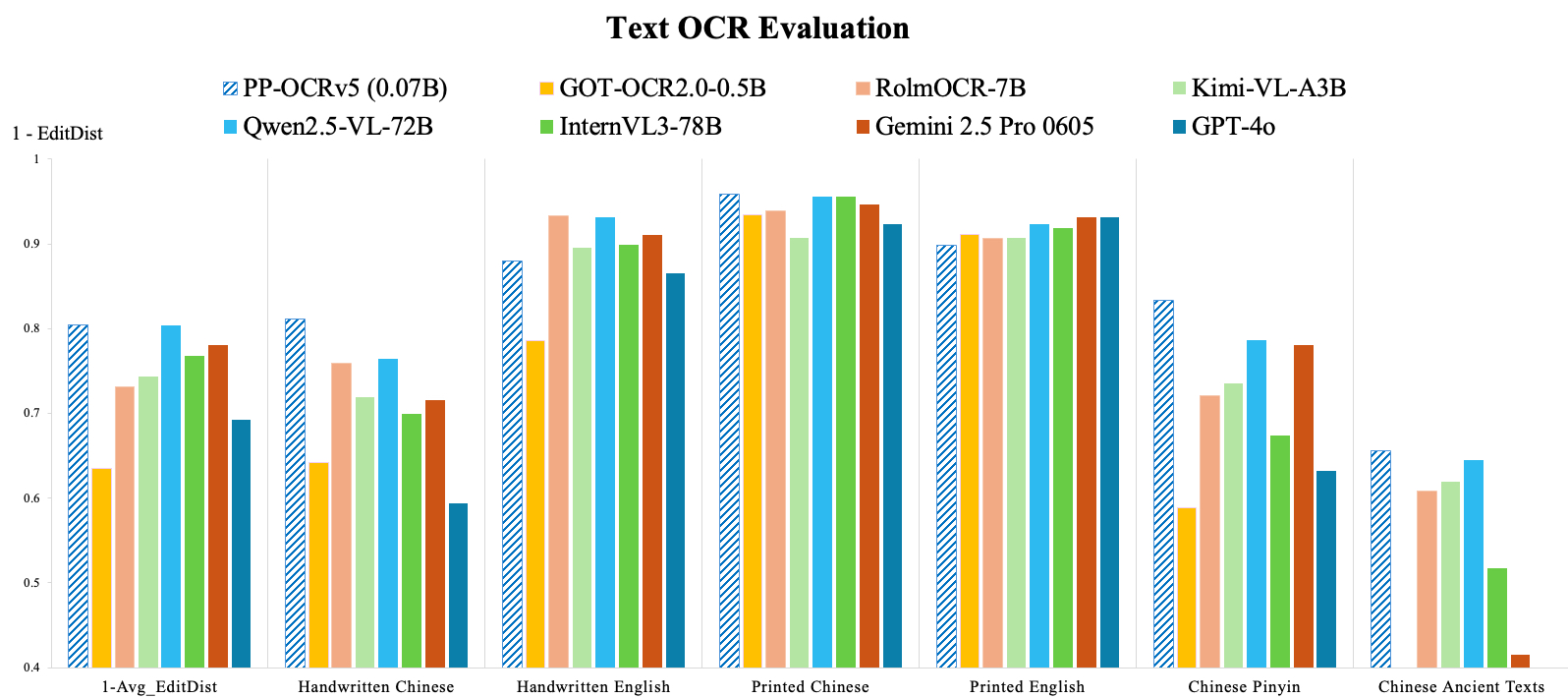}
\caption{
    \centering
    Text OCR evaluation. The term "1-EditDist" refers to \(1 - \text{Edit Distance}\), a higher value of this metric indicates superior performance.
    }
\label{fig:ocr_benchmark}
\end{figure}

We evaluated the OCR capabilities across 17 different scenarios, including handwritten Chinese, handwritten English, printed Chinese, printed English, Chinese Pinyin, Japanese, Chinese ancient texts, traditional Chinese, common, blurred, rotated, Greek characters, emojis, tables, artistic fonts, special symbols, and deformed scenes. These diverse scenarios allow us to comprehensively assess the performance and adaptability of the system. Based on the OmniDocBench OCR text evaluation standards, we conducted extensive testing on mainstream OCR methods and multimodal large models. Figure~\ref{fig:ocr_benchmark} presents the evaluation results, using 1-edit distance as the metric, and lists the average metrics across all scenarios as well as specific performance metric for key scenarios, including handwritten and printed Chinese and English, Chinese Pinyin, and Chinese ancient texts.

The results indicate that the lightweight PP-OCRv5 ranks first in terms of the average 1-edit distance across all scenarios, surpassing all multimodal large models such as GOT-OCR2.0-0.5B~\citep{wei2024general}, RolmOCR-7B~\citep{RolmOCR}, Qwen2.5-VL-72B~\citep{qwen2.5}, InternVL3-78B~\citep{chen2024expanding}, Gemini 2.5 pro 0605\footnote{\label{gemini2.5pro}\url{https://deepmind.google/models/gemini/pro/}}, and GPT-4o\footnote{\label{gpt4o}\url{https://openai.com/index/hello-gpt-4o/}}. In Chinese scenarios, whether handwritten or printed, PP-OCRv5 significantly outperforms other methods. Although slightly inferior to multimodal solutions in handwritten English recognition, it is noteworthy that our model has only 0.07~B parameters. In the Chinese Pinyin and Chinese ancient text scenarios, PP-OCRv5 has achieved significant advantages. These results demonstrate that lightweight models specifically designed for OCR tasks can match or even exceed the accuracy of large-scale multimodal models. Simultaneously, they offer substantially reduced computational and storage requirements, significantly enhancing inference efficiency and deployment flexibility, and thereby delivering critical advantages for industrial applications and mobile device deployment.

\subsection{PP-StructureV3} 

\begin{figure}[!h]
\centering
\includegraphics[width=0.998\textwidth]{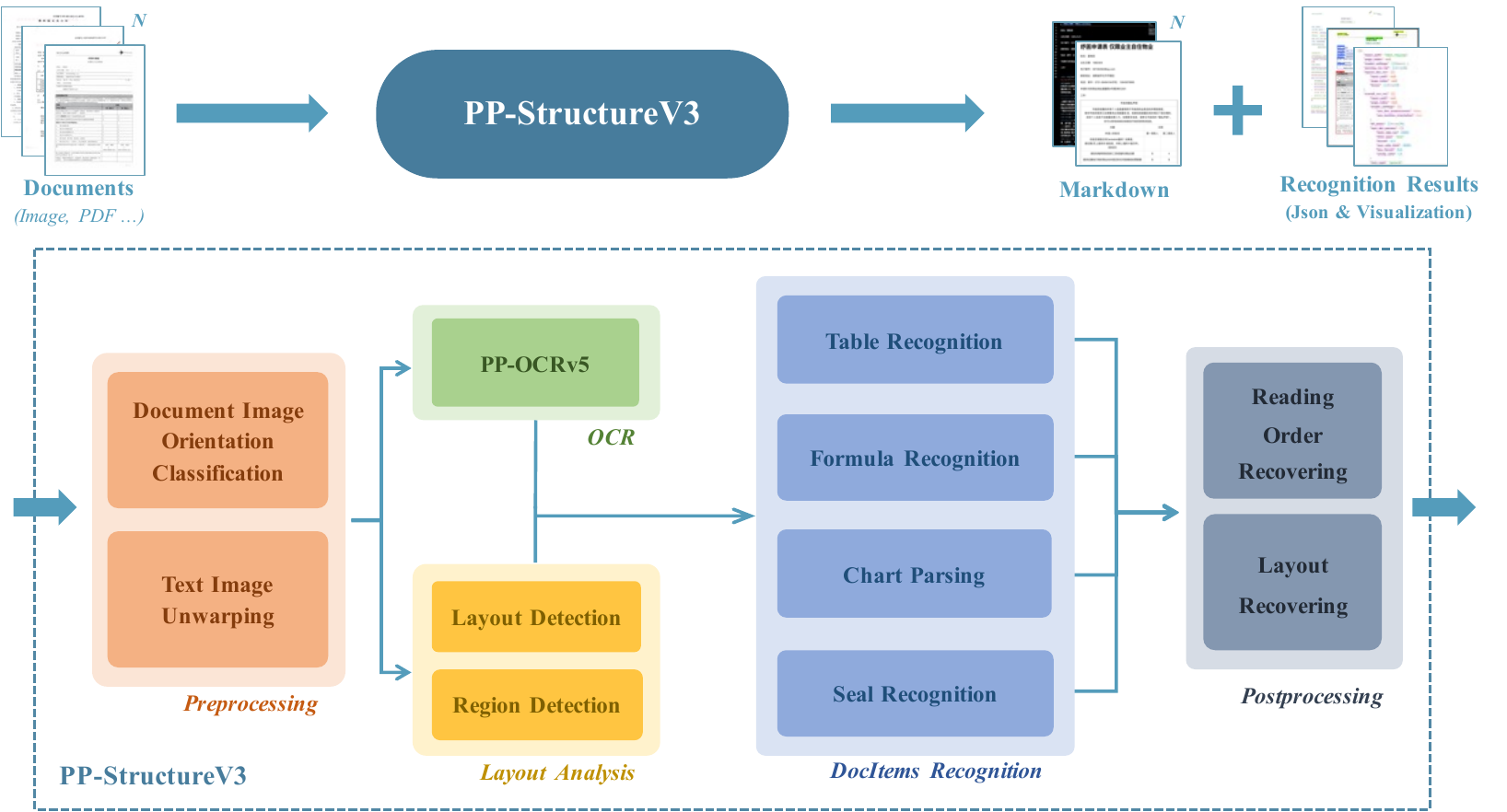}
\caption{
    \centering
    Pipeline of PP-StructureV3. The pipeline includes Preprocessing, OCRv5, Layout Analysis and Document Items Recognition and Postprocessing. It effectively parses content from images and outputs it as structured data.
}
\label{fig:pp_structurev3_framwork}
\end{figure}

PP-StructureV3 is a multi-model pipeline system developed for document image parsing tasks, it can accurately and efficiently convert document images or PDF files into structured JSON files and Markdown files. As illustrated in the algorithm framework in Figure~\ref{fig:pp_structurev3_framwork}, the system primarily consists of five modules: preprocessing, OCR, layout analysis, document item recognition, and postprocessing.

\begin{table}[!h]
    \centering
    \begin{threeparttable}
    \begin{tabular}{c|c|cc}
        \toprule
        \multirow{2}{*}{Method Type}    & \multirow{2}{*}{Methods}      & \multicolumn{2}{c}{Edit $\downarrow$} \\
                                                                          \cmidrule{3-4}              
                                        &                               & EN             & ZH             \\
        \midrule
        \multirow{9}{*}{Pipeline Tools} & PP-StructureV3                                        & \textbf{0.145} & \textbf{0.206} \\
                                        & MinerU-1.3.11~\citep{wang2024mineru}                    & 0.166          & 0.310          \\
                                        & MinerU-0.9.3~\citep{wang2024mineru}                     & 0.150          & 0.357          \\
                                        & Mathpix\tnote{1}                & 0.191          & 0.365          \\
                                        & Pix2Text-1.1.2.3~\citep{Pix2Text}                       & 0.320          & 0.528          \\
                                        & Marker-1.2.3~\citep{marker}                             & 0.336          & 0.556          \\
                                        & Unstructured-0.17.2~\citep{unstructured}                & 0.586          & 0.716          \\
                                        & OpenParse-0.7.0~\citep{open-parse}                      & 0.646          & 0.814          \\
                                        & Docling-2.14.0~\citep{Docling_Team_Docling}             & 0.589          & 0.909          \\
        \midrule
        \multirow{5}{*}{Expert VLMs}    & GOT-OCR2.0~\citep{wei2024general}                          & 0.287          & 0.411          \\
                                        & Mistral OCR\tnote{2}             & 0.268          & 0.439          \\
                                        & OLMOCR-sglang~\citep{poznanski2025olmocr}               & 0.326          & 0.469          \\
                                        & SmolDocling-256M\_transformer~\citep{nassar2025smoldocling} & 0.493          & 0.816          \\
                                        & Nougat~\citep{blecher2023nougat}                        & 0.452          & 0.973          \\
        \midrule
        \multirow{5}{*}{General VLMs}   & Gemini2.5-Pro\tnote{3}        & 0.148          & 0.212          \\
                                        & Gemini2.0-flash\tnote{4}  & 0.191          & 0.264          \\
                                        & Qwen2.5-VL-72B~\citep{qwen2.5}                                                      & 0.214          & 0.261          \\
                                        & GPT-4o\tnote{5}    & 0.233          & 0.399          \\
                                        & InternVL2-76B~\citep{chen2024expanding}                                             & 0.440          & 0.443          \\
        \bottomrule
    \end{tabular}
    \begin{tablenotes}
        \footnotesize
        \item[1] \url{https://mathpix.com/}
        \item[2] \url{https://mistral.ai/}
        \item[3] \url{https://deepmind.google/models/gemini/pro/}
        \item[4] \url{https://deepmind.google/models/gemini/flash/}
        \item[5] \url{https://openai.com/index/hello-gpt-4o/}
    \end{tablenotes}
    \caption{Comprehensive evaluation of document parsing methods on OmniDocBench.}
    \label{tab:pp-structurev3-omnidocbench}
    \end{threeparttable}
\end{table}

1. \textbf{Preprocessing}: Similar to PP-OCRv5 (see Section~\ref{sec:pp-ocrv5}), this module comprises a document image orientation classification model based on PP-LCNet and a text image unwarping model based on UVDoc. It is primarily designed to address issues related to low-quality document images, such as rotation and distortion.

2. \textbf{OCR}: This module employs PP-OCRv5 (see Section~\ref{sec:pp-ocrv5}) with preprocessing disabled to detect and recognize all textual content within document images. Compared to PP-OCRv4, PP-OCRv5 achieves significant performance improvements via optimizations in network architecture, training strategies (such as knowledge distillation), and enhancements to the training dataset. Notably, it demonstrates substantial improvements in detection and recognition for challenging scenarios, including vertical text layouts, handwritten text, and rare Chinese characters.

3. \textbf{Layout Analysis}: This module incorporates two models: a layout detection model and a region detection model. The layout detection model, PP-DocLayout-plus, is an optimized version of the PP-DocLayout\citep{sun2025pp}. It significantly enhances layout detection performance for complex documents, such as multi-column magazines and newspapers, reports with multiple tables, exams, handwritten documents, Japanese and vertically oriented layouts documents. The newly proposed layout region detection model addresses the problem of multiple articles appearing on a single layout page. For example, a single newspaper page often contains several distinct articles. Using only the layout region detection model makes it challenging to correctly associate elements with their respective articles, leading to errors in reading order recovery. By introducing the layout region detection model, elements can be accurately assigned to their corresponding articles.

4. \textbf{Document Items Recognition}: Based on predictions from the layout detection models, the content of each page element is recognized using appropriate methods, including the table recognition solution PP-TableMagic, the formula recognition model PP-FormulaNet\_plus, the chart parsing model PP-Chart2Table, and seal recognition with PP-OCRv4\_seal.

\begin{itemize}
    \item \textbf{Table Recognition}: PP-TableMagic is a comprehensive table recognition system composed of several specialized models. It includes a table orientation classification model and a frame type classification model, which determine the rotation and framing style of the table, respectively, guiding the selection of the appropriate recognition method. The cell detection model, based on object detection algorithms, accurately locates individual table cells. Additionally, the structure recognition model outputs the table’s structure in HTML format.
    \item \textbf{Formula Recognition}: This model is an enhanced version of PP-FormulaNet~\citep{liu2025formula}, capable of recognizing images containing formulas cropped from full document images and generating the corresponding LaTeX code. To address the recognition of complex multi-line formulas, the token length was increased to 2560, and the training dataset was expanded to include more complex formulas. Additionally, to handle formulas containing Chinese characters, a large volume of relevant data was mined for training.
    \item \textbf{Chart Parsing}: PP-Chart2Table is a lightweight, end-to-end vision-language model designed to accurately extract data from various types of chart images, such as histograms, line charts, and pie charts, and convert the extracted information into tables represented in markdown format. It genuinely understands and retrieves chart data through an innovative Shuffled Chart Data Retrieval task and meticulous token masking. Its performance is further boosted by a sophisticated data synthesis pipeline that generates diverse, high-quality training data using RAG with high-quality seeds and LLM persona design. A two-stage LLM distillation process, leveraging large volumes of unlabeled out-of-distribution (OOD) data, enhances the model's adaptability and generalization to real-world scenarios.
    \item \textbf{Seal Recognition}: PP-OCRv4\_seal is a system specifically tailored for the recognition of oval, round, and other types of seals. It incorporates a curved text detection model, which can accurately detect and rectify bent text, as well as a general-purpose text recognition model.
\end{itemize}

5. \textbf{Post-processing Module}: Upon completion of the above steps, the positions and corresponding content of each element in the document are obtained. The post-processing module then reconstructs the relationships among elements, such as linking figures and tables with their captions and recovering the correct reading order. PP-StructureV3 improves upon the X-Y Cut~\citep{ha1995recursive}, significantly enhancing the reconstruction of reading order in complex layouts, including magazines, newspapers, and vertically typeset documents.

To evaluate the performance of PP-StructureV3, we conducted experiments using the OmniDocBench benchmark, with the results presented in Table~\ref{tab:pp-structurev3-omnidocbench}. As shown in the table, PP-StructureV3 demonstrates exceptional performance in Chinese and English document parsing, establishing itself as the current SOTA. It not only significantly outperforms other pipeline-based tools, but also shows strong competitiveness when compared to the most popular expert VLMs and general VLMs.

\subsection{PP-ChatOCRv4}

\begin{figure}[!h]
\centering
\includegraphics[width=0.998\textwidth]{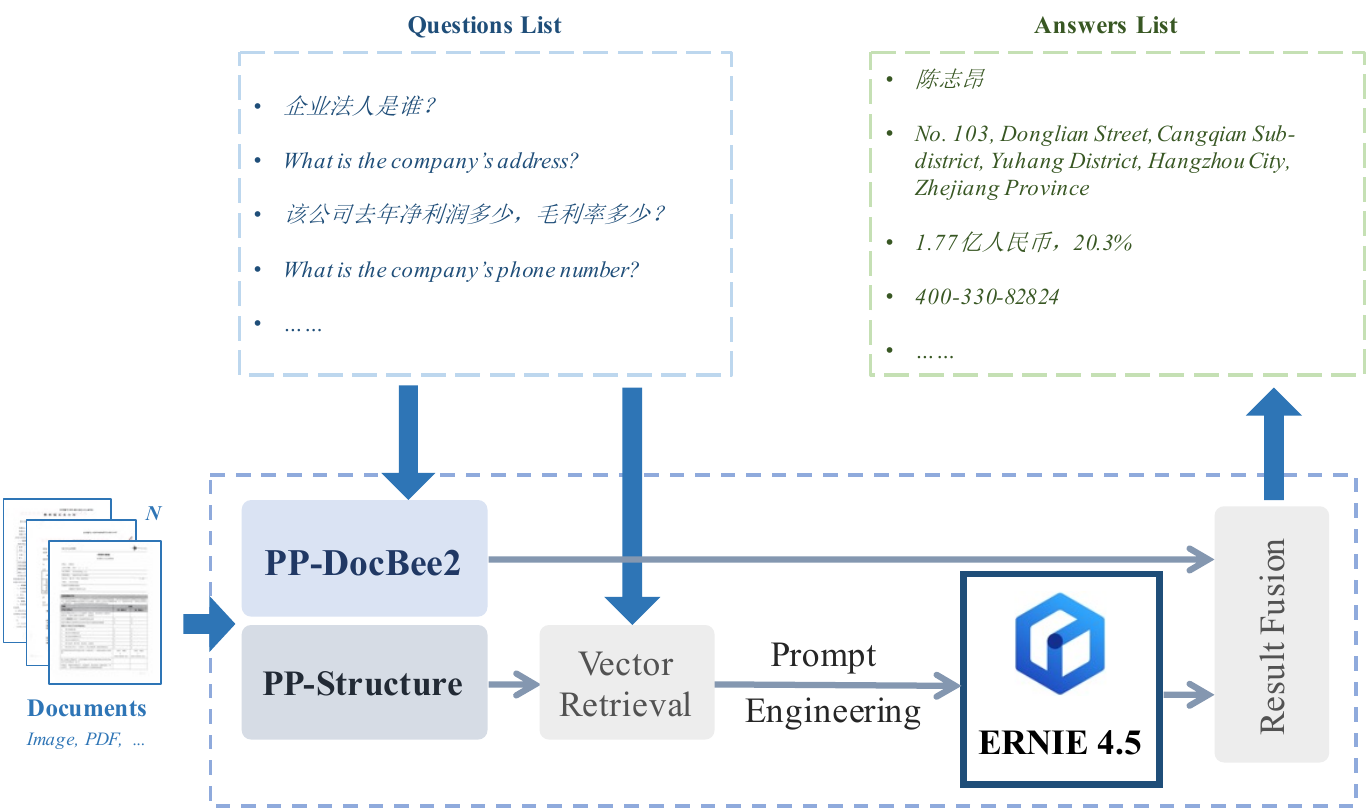}
\caption{
    \centering
    Pipeline of PP-ChatOCRv4.
}
\label{fig:pp_charocrv4_framwork}
\end{figure}

PP-ChatOCRv4 is an advanced key information extraction solution for document images, leveraging LLMs, VLMs, and OCR technologies to enable robust key information extraction in challenging scenarios such as complex layouts, multi-page PDFs, rare characters, intricate table structures, and documents containing seals. As illustrated in Figure~\ref{fig:pp_charocrv4_framwork}, the overall workflow of PP-ChatOCRv4 comprises the following components: the layout analysis module PP-Structure~\citep{li2022ppstructurev2}, the vector retrieval module, the large language model ERNIE 4.5, the document-oriented vision-language model PP-DocBee2~\citep{ni2025pp}, and the result fusion module.

1. \textbf{PP-Structure}: PP-Structure serves as the document image parsing module. It is built upon multiple specialized models, including layout detection, text line detection, text recognition, and table structure recognition models. By leveraging these models, PP-Structure is able to parse various elements from document images, thereby generating a structured, text-based representation of the document content.

2. \textbf{Vector Retrieval Module}: A feature vector database is constructed using the textual content parsed from document images. During key information extraction, RAG technology is initially employed to efficiently identify and extract critical information from lengthy and redundant texts. By leveraging RAG, the efficiency and accuracy of information retrieval are significantly enhanced.

3. \textbf{Large Language Model}: The framework supports any large language model (LLM); for example, we currently use ERNIE-4.5-300B-A47B, the latest LLM released by Baidu, to extract information based on carefully designed prompts. These prompts are crafted to seamlessly integrate retrieved textual information with user queries, thereby enhancing both the efficiency and accuracy of the results.

4. \textbf{PP-DocBee2}: The PP-DocBee2 is a novel multimodal large language model with 3 billion parameters designed for end-to-end document image understanding developed by us. It is capable of directly extracting text-based answers from document images using prompts constructed from the given questions.

5. \textbf{Result Fusion}: Extraction results from both text-based and image-based approaches are fused to produce the final output.

To evaluate the performance of PP-ChatOCRv4 and other methods, we designed an end-to-end evaluation pipeline using a custom multi-scenario benchmark dataset comprising 638 document images. These images span a wide range of scenarios, including financial reports, research papers, contracts, manuals, regulations, as well as humanities and science papers. Each document image is accompanied by several questions and corresponding answers, resulting in a total of 1,196 question-answer pairs. The results are presented in Table~\ref{tab:pp-charocrv4-recall}.

\begin{table}[!h]
    \centering
    \begin{tabular}{c|c}
        \toprule
        Methods         &  Recall@1             \\
        \midrule
        GPT-4o          &  63.47\%              \\
        PP-ChatOCRv3    &  70.08\%              \\
        Qwen2.5-VL-72B  &  80.26\%              \\
        PP-ChatOCRv4    &  \textbf{85.55}\%     \\
        \bottomrule
    \end{tabular}
    \caption{Comprehensive evaluation of various solutions on a custom multi-scenario benchmark: overall recall@1 performance based on ground truth comparison.}
    \label{tab:pp-charocrv4-recall}
\end{table}

\section{Codebase Architecture Design}

\subsection{Overall Architecture}

\begin{figure}[h]
\centering
\includegraphics[width=0.95\textwidth]{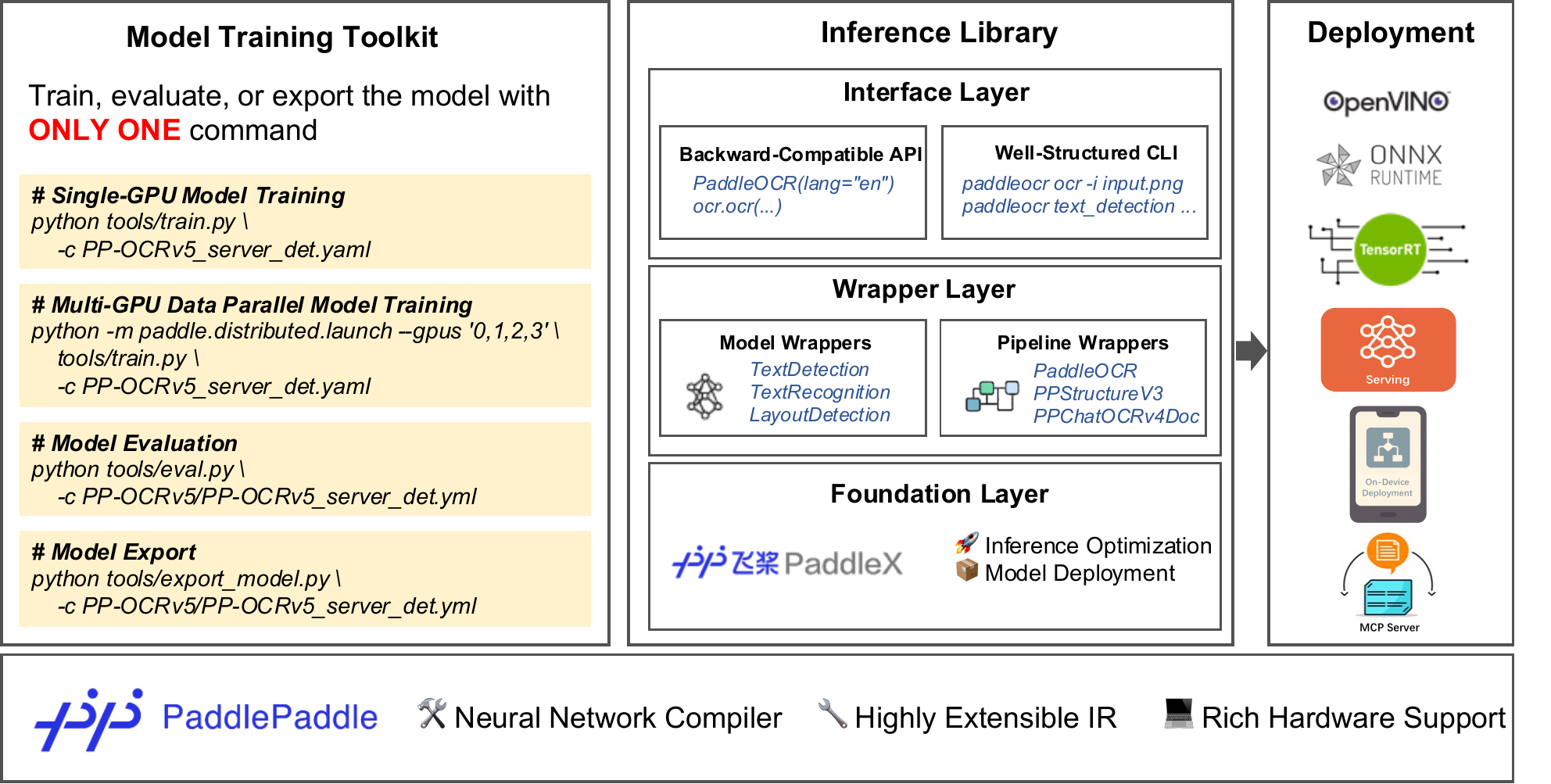}
\caption{
    \centering
    Overall Architecture of the PaddleOCR~3.0 Codebase.
}
\label{fig:codebase_overall_arch}
\end{figure}

The overall architecture of the PaddleOCR~3.0 codebase is illustrated in Figure~\ref{fig:codebase_overall_arch}. At its foundation, PaddleOCR~3.0 is built upon the PaddlePaddle framework~\citep{ma2019paddlepaddle}, which incorporates a neural network compiler for performance optimization, supplies a highly extensible intermediate representation (IR), and ensures broad hardware compatibility. Building on this foundation, PaddleOCR~3.0 is structured around two core components: a model training toolkit and an inference library. The inference library offers flexible integration paths that naturally extend to deployment. The deployment capabilities will be introduced in Section~\ref{sec:deployment}.

The training toolkit provides a comprehensive suite of utilities that support the complete training pipeline for various models, including text detection models, text recognition models, etc. It also facilitates the conversion of trained models from dynamic graph format to static graph format, thereby enhancing their suitability for inference and deployment in production environments. Users can execute Python scripts with a single command to perform tasks such as model training, evaluation, and export. Additionally, various parameters can be configured to meet different requirements, such as specifying the path to a pre-trained model or using a custom dataset directory.

Complementing this, the inference library is designed to be lightweight and highly efficient. It supports loading both officially released inference models and custom models trained by users. The library enables inference across eight end-to-end model pipelines and can be readily integrated into real-world applications. We will elaborate the design of the inference library in the next subsection. The inference library serves as the foundation for downstream deployment capabilities, including high-performance inference across various frameworks, deploying the model pipeline as a service, deploying it to mobile devices, and running it via a Model Context Protocol (MCP) server.

\subsection{Inference Library Design}

In this subsection, we present the rationale behind the design and structural organization of the PaddleOCR~3.0 inference library.

Let us start with the defects of the inference library in PaddleOCR~2.x:

\begin{itemize}
    \item All parameters of the PaddleOCR~2.x CLI exist in the same namespace. This is not extensible, as each new feature required manually adding more arguments to an increasingly bloated global parameter list, making it difficult to maintain and scale.
    
    \item In PaddleOCR~2.x, the parameters are only configurable through function arguments. Such an approach hinders reproducibility and portability, especially in scenarios where configurations need to be shared or version-controlled.
    
    \item PaddleOCR~2.x lacks clear separation of concerns---the boundary between the model development toolkit (primarily designed for training) and the inference library was not clearly defined. Instead, the inference library was built directly on top of the model development toolkit inference scripts, which introduced two entry points for the inference functionality, potentially causing confusion for users. Additionally, the inference library was constructed based on assumptions about what the entry points should be, which broke modularity and maintainability.
\end{itemize}

To address these issues, we upgraded the inference library on top of the PaddleX~3.0 toolkit\footnote{\label{paddlex3.0}\url{https://github.com/PaddlePaddle/PaddleX/tree/release/3.0}}, which provides extensive inference optimization and deployment features. The backward compatibility was also considered, which minimizes migration effort for users transitioning from PaddleOCR~2.x. The new inference library consists of three layers, as illustrated in Figure~\ref{fig:codebase_overall_arch}.

\begin{itemize}
    \item \textbf{Interface Layer}: The library offers both a Python API and a CLI for user interaction. In PaddleOCR~3.0, all OCR tasks are accessed through a consistent and unified Python API. To facilitate a smooth transition for existing users, the API preserves backward compatibility for key methods and parameters. The CLI has been completely redesigned compared to PaddleOCR~2.x, introducing subcommands that clearly distinguish between different tasks and thereby providing a cleaner, more intuitive user experience.
    
    \item \textbf{Wrapper Layer}: This layer offers Pythonic wrappers for core PaddleX components, including models and pipelines. These wrappers deliver unified interfaces and flexible configuration management. In addition to maintaining the backward compatibility in the argument-based approach previously preferred in PaddleOCR~2.x, the PaddleX-style configuration file-based approach is also supported, which allows configurations to be stored and reused in a portable and reproducible manner.
    
    \item \textbf{Foundation Layer}: At the foundation lies the PaddleX~3.0 toolkit, which forms the core of PaddleOCR~3.0. It offers powerful features for inference optimization and model deployment, which are fully integrated into PaddleOCR~3.0. Transferring the basis from PaddleOCR scripts to PaddleX ensures the separation of roles of the model training toolkit and the inference library, eliminating redundant entry points and clarifying functional boundaries. This decoupling allows each component to evolve independently, reduces user confusion, and lays the foundation for a more robust and maintainable system design.
\end{itemize}

This layered architecture ensures that higher-level components depend only on lower-level abstractions, promoting loose coupling, modularity, and ease of maintenance.

\section{Deployment}\label{sec:deployment}

\begin{figure}[h]
\centering
\includegraphics[width=\textwidth]{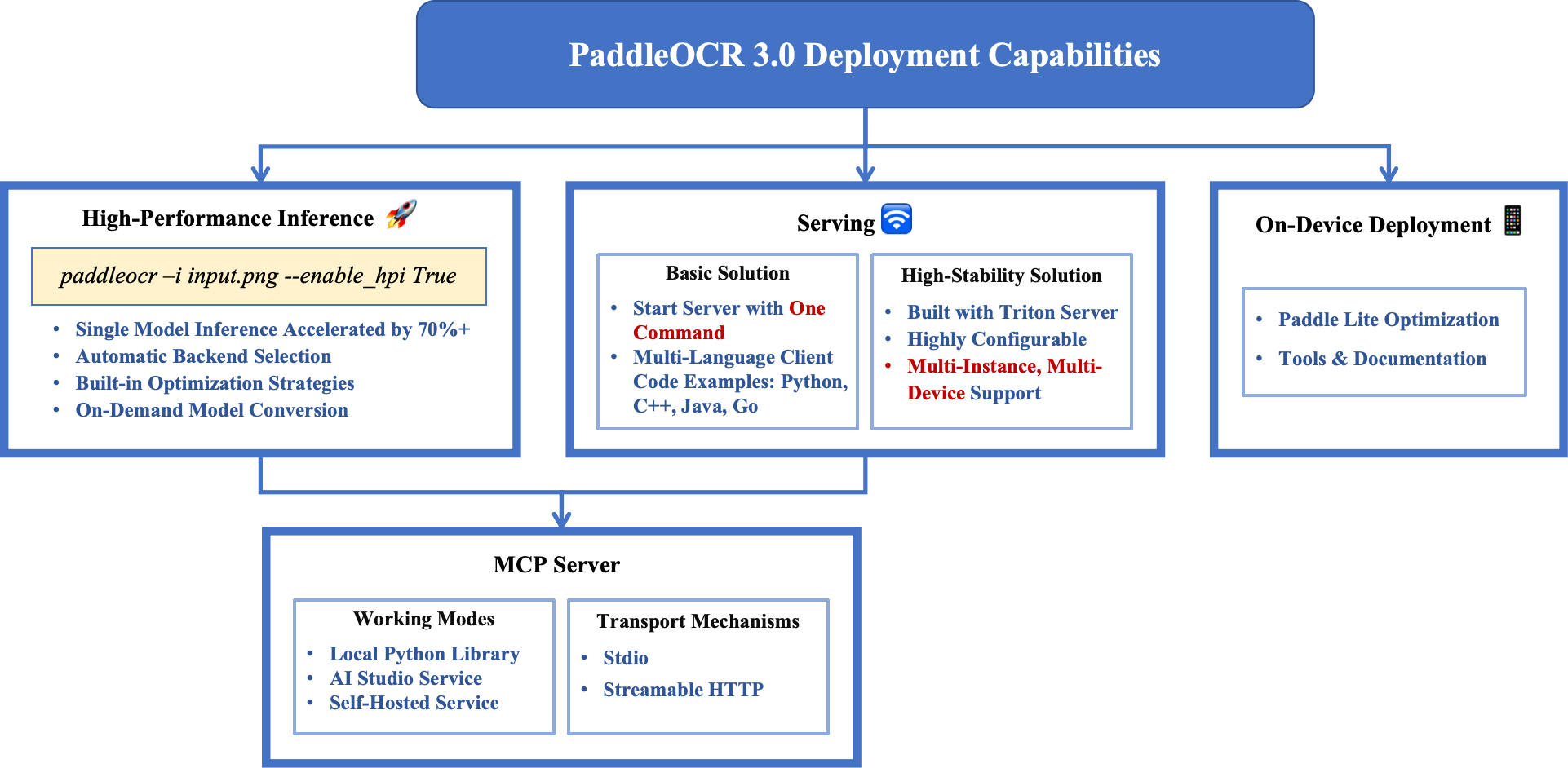}
\caption{
    \centering
    Overview of the deployment capabilities of PaddleOCR~3.0. PaddleOCR~3.0 provides high-performance inference, serving, and on-device deployment capabilities. Additionally, it enables users to easily deploy an MCP server based on PaddleOCR.
}
\label{fig:deployment_overview}
\end{figure}

An overview of the deployment capabilities of PaddleOCR~3.0 is depicted in Figure~\ref{fig:deployment_overview}. To support a wide range of application scenarios, PaddleOCR~3.0 offers flexible and comprehensive deployment options, including high-performance inference, serving, and on-device deployment. In real-world production environments, OCR-related systems are often subject to constraints beyond recognition accuracy, such as latency, throughput, and hardware compatibility. PaddleOCR addresses these requirements by providing configurable deployment tools that simplify integration across various platforms. In addition, to facilitate integration with LLM applications, PaddleOCR provides an MCP server, which allows users to leverage high-performance inference pipelines or pipeline servers.

\subsection{High-Performance Inference}

PaddleOCR~3.0 provides the high-performance inference feature that enables users to optimize runtime performance without the need to manually tune low-level configurations. High-performance inference provides notable acceleration for some key models. For instance, on NVIDIA Tesla T4 devices, enabling high-performance inference reduces the single-model inference latency of PP-OCRv5\_mobile\_rec by 73.1\% and that of PP-OCRv5\_mobile\_det by 40.4\%. The key features of PaddleOCR~3.0's high-performance inference capability include:

\begin{itemize}
  \item Automatic selection of appropriate inference backends based on the runtime environment and model characteristics, including support for Paddle Inference, OpenVINO~\citep{openvino}, \mbox{ONNX Runtime}~\citep{onnxruntime}, and TensorRT~\citep{tensorrt}.
  \item Built-in optimization strategies such as multi-threading and FP16 inference to better utilize hardware resources.
  \item On-demand model conversion from PaddlePaddle static graphs to ONNX format to enable acceleration on compatible inference engines.
\end{itemize}

Users can easily achieve inference acceleration by enabling the \texttt{enable\_hpi} switch, while all underlying optimization details are managed by PaddleOCR. For advanced needs, PaddleOCR also supports fine-grained tuning of high-performance inference configurations in pipelines by passing Python/command line parameters or modifying configuration files.

\subsection{Serving}

PaddleOCR~3.0 supports pipeline serving for building scalable and production-ready OCR-related services. Two solutions are provided:

\begin{itemize}
  \item \textbf{Basic Serving}: A lightweight solution based on FastAPI~\citep{fastapi} with minimal setup, suitable for rapid validation and scenarios with low concurrency requirements. For this solution, users can run any pipeline as a service with a single command via the CLI. For the client-side code, PaddleOCR~3.0 provides rich calling examples in seven programming languages: Python, C++, Java, Go, C\#, Node.js, and PHP. Users can refer to these examples to quickly integrate the service capabilities into their own applications.
  \item \textbf{High-Stability Serving}: A more robust option built on NVIDIA Triton Inference Server~\citep{tritoninference}, which supports more advanced deployment configurations. This solution is suitable for scenarios with higher requirements for stability and performance. For example, the server can be configured to run multiple instances across multiple GPUs to fully utilize the available computing resources.
\end{itemize}

Both solutions share similar interfaces. Users can start with the basic solution for rapid validation, and then decide whether to adopt the more complex high-stability solution according to their needs, usually without significant migration costs.

\subsection{On-Device Deployment}

To support the deployment on resource-constrained devices, PaddleOCR~3.0 enables deployment of PP-OCR models on mobile platforms. It provides supporting tools and documentation for model optimization and integration with the Paddle-Lite\footnote{\url{https://github.com/PaddlePaddle/Paddle-Lite}}, runtime, making it feasible to run OCR tasks efficiently on mobile devices.

\subsection{MCP Server}

PaddleOCR~3.0 provides a lightweight MCP server, enabling smooth integration of PaddleOCR’s core capabilities into any MCP-compatible host. Both the OCR and PP-StructureV3 pipelines are currently accessible as tools via the MCP server.

Built on top of the PaddleOCR inference library, the MCP server supports various inference and deployment methods provided by PaddleOCR. At present, it can operate in one of three working modes:

\begin{itemize}
    \item \textbf{Local:} Runs the PaddleOCR pipeline directly on the local machine using the installed Python library. This mode is suitable for offline usage and situations with strict data privacy requirements. High-performance inference can be activated to accelerate the inference process.
    \item \textbf{AI Studio:} Utilizes cloud services hosted by the PaddlePaddle AI Studio community~\citep{aistudio}. This mode is ideal for quickly trying out features, validating solutions, and for no-code development scenarios. 
    \item \textbf{Self-Hosted:} Connects to a user-hosted PaddleOCR service. This mode offers the advantages of pipeline serving and high flexibility, making it well-suited for scenarios requiring custom service configurations.
\end{itemize}

Regardless of the selected working mode, setting up the PaddleOCR MCP server is straightforward for users. Example configuration files are included in the appendix~\ref{sec:appendix} for reference. Additionally, the PaddleOCR MCP server supports both stdio and Streamable HTTP transport mechanisms, offering flexibility for a wide range of deployment scenarios. With its adaptable architecture and support for multiple deployment modes, the PaddleOCR MCP server can effectively address diverse real-world application needs, delivering robust and scalable solutions for both individual developers and enterprise users.

\section{Conclusion}

PaddleOCR has been dedicated to the field of OCR and document parsing for many years, aiming to provide more valuable technical solutions. PaddleOCR 3.0 is a milestone upgrade, with technologies like PP-OCRv5, PP-StructureV3, and PP-ChatOCRv4 set to play a significant role in the era of large-scale models. Moving forward, we will continue to expand our models, including the upcoming release of multilingual text recognition models, multimodal OCR, and document parsing models. If you find PaddleOCR 3.0 useful or wish to use it in your projects, please kindly cite this technical report.

\bibliography{main}

\newpage
\appendix

\section*{Appendix}
\label{sec:appendix}

\section{Acknowledgments}
We gratefully acknowledge all individuals who supported this work through their invaluable contributions to data construction, deployment, testing, project maintenance, product development, online demo creation, and operations. Their dedication and efforts have played a crucial role in the successful advancement and ongoing improvement of this project.

\setlength{\parskip}{0pt} % 让段落之间没有额外空隙
\setlength{\itemsep}{0pt} % 如果用itemize
\setlength{\parsep}{0pt}  % 控制段落间距
\setlength{\parindent}{0pt} % 首段取消空格
\begin{multicols}{4}
Baoku Yu\\
Chang Xu\\
Chao Han\\
Chunli Xie\\
Guanzhong Wang\\
Haitao Yu\\
Hengxin Chen\\
Hong Cheng\\
Jiahao Bai\\
Jiahua Wang\\
Jianying Qu\\
Jiaxin Sui\\
Jinghui Duan\\
JingsongLiu \\
Mengmeng Guo\\
Min Zhuang\\
Runlong Li\\
Shengjian Guo\\
Siyu Cheng\\
Suyin Liang\\
Tao Luo \\
Tianyu Zheng\\
Xiaolong Ma\\
Xin Li\\
Xin Wang \\
Xinran Liu\\
Yimin Gao\\
Yiqiao Zhou\\
Ye Han\\
Yongkun Du \\
Zewu Wu\\
Zeyu Luo\\
Zhe Wang\\
Zhongkai Sun\\

\end{multicols}

We would also like to acknowledge the invaluable contributions of open-source developers on GitHub, including but not limited to \href{https://github.com/timminator}{@timminator}, \href{https://github.com/ackinc}{@ackinc}, \href{https://github.com/Appla}{@Appla}, \href{https://github.com/co63oc}{@co63oc}, \href{https://github.com/jk4e}{@jk4e}, and many others whose work has inspired and supported our project.

\begin{figure}[htbp]
    \centering
    \includegraphics[width=0.8\linewidth]{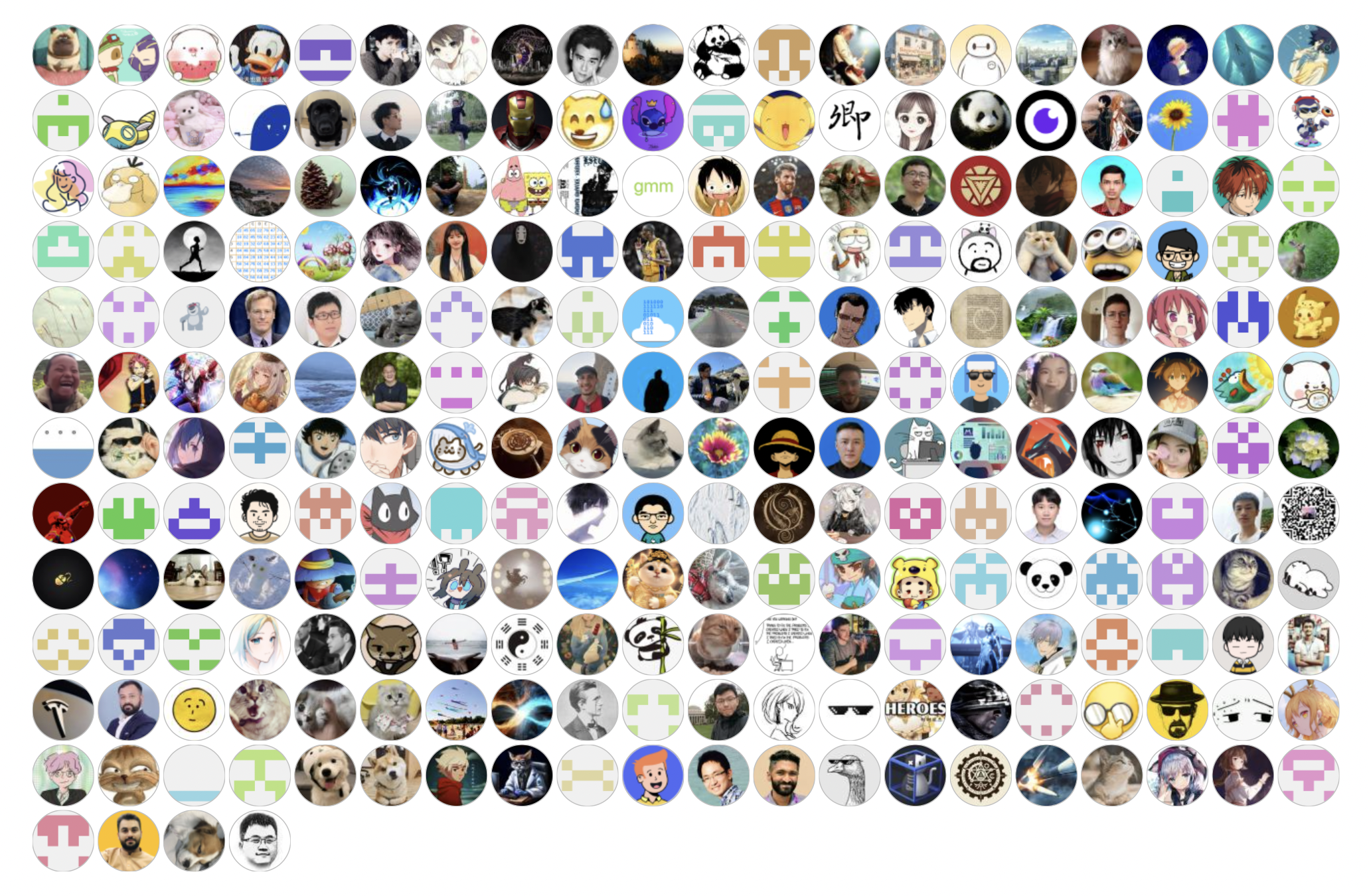}
    \caption{Avatars of long-term contributors to the PaddleOCR project.}
    \label{fig:contributors}
\end{figure}

\vspace{1em} 
Finally, we express our sincere gratitude to all project contributors for their long-term support and valuable input. Their efforts have greatly advanced the development and improvement of this project. Figure~\ref{fig:contributors} shows the avatars of these contributors, as collected from the PaddleOCR GitHub repository.

\newpage
\section{Usage of command and API details}

To use PaddleOCR 3.0, you can simply install the paddleocr package from PyPI. PaddleOCR 3.0 provides command-line interface (CLI) and Python API for users to use conveniently. 

\subsection{Run inference by CLI}

We provide convenient CLI methods for users to quickly experience the capabilities of PP-OCRv5, PP-StructureV3, and PP-ChatOCRv4, as follows:

\begin{lstlisting}[language=bash]
# Run PP-OCRv5 inference
paddleocr ocr -i test.png \
        --use_doc_orientation_classify False \
        --use_doc_unwarping False \
        --use_textline_orientation False 

# Run PP-StructureV3 inference
paddleocr pp_structurev3 -i test.png \
        --use_doc_orientation_classify False \
        --use_doc_unwarping False

# Get the Qianfan API Key at first, then run PP-ChatOCRv4
paddleocr pp_chatocrv4_doc -i test.png \
        -k number \
        --qianfan_api_key your_api_key \
        --use_doc_orientation_classify False \
        --use_doc_unwarping False  
\end{lstlisting}

\subsection{Run inference by Python API}

We also provide a clean interface to facilitate users in using and integrating it into their own projects.

\textbf{1. PP-OCRv5 Example}

\begin{lstlisting}[language=Python]
# Initialize PaddleOCR instance
from paddleocr import PaddleOCR
ocr = PaddleOCR(
    use_doc_orientation_classify=False,
    use_doc_unwarping=False,
    use_textline_orientation=False)

# Run OCR inference on a sample image 
result = ocr.predict(input="test.png")

# Visualize the results and save the JSON results
for res in result:
    res.print()
    res.save_to_img("output")
    res.save_to_json("output")
\end{lstlisting}

\textbf{2. PP-StructureV3 Example}

\begin{lstlisting}[language=Python]
# Initialize PPStructureV3 instance
from paddleocr import PPStructureV3

pipeline = PPStructureV3(
    use_doc_orientation_classify=False,
    use_doc_unwarping=False)

# Run PPStructureV3 inference 
output = pipeline.predict(input="test.png")

# Visualize the results and save the JSON results
for res in output:
    res.print() 
    res.save_to_json(save_path="output") 
    res.save_to_markdown(save_path="output")    
\end{lstlisting}

\textbf{3. PP-ChatOCRv4 Example}

\begin{lstlisting}[language=Python]
from paddleocr import PPChatOCRv4Doc

chat_bot_config = {
    "module_name": "chat_bot",
    "model_name": "xxx",
    "base_url": "https://qianfan.baidubce.com/v2",
    "api_type": "openai",
    "api_key": "api_key",  # your api_key
}

retriever_config = {
    "module_name": "retriever",
    "model_name": "embedding-v1",
    "base_url": "https://qianfan.baidubce.com/v2",
    "api_type": "qianfan",
    "api_key": "api_key",  # your api_key
}

# Initialize PPChatOCRv4Doc instance
pipeline = PPChatOCRv4Doc(
    use_doc_orientation_classify=False,
    use_doc_unwarping=False)

visual_predict_res = pipeline.visual_predict(
    input="test.png",
    use_common_ocr=True,
    use_seal_recognition=True,
    use_table_recognition=True)

mllm_predict_info = None
use_mllm = False 
visual_info_list = []

for res in visual_predict_res:
    visual_info_list.append(res["visual_info"])
    layout_parsing_result = res["layout_parsing_result"]

vector_info = pipeline.build_vector(
    visual_info_list, 
    flag_save_bytes_vector=True, 
    retriever_config=retriever_config)
    
chat_result = pipeline.chat(
    key_list=["number of people"],
    visual_info=visual_info_list,
    vector_info=vector_info,
    mllm_predict_info=mllm_predict_info,
    chat_bot_config=chat_bot_config,
    retriever_config=retriever_config)
print(chat_result) 
\end{lstlisting}

\section{More details on MCP host configuration}

Here are several example configurations for the MCP host in Claude for Desktop, illustrating how to connect to a PaddleOCR MCP server. Each configurable parameter may be specified either via environment variables (as shown in the \texttt{env} field) or via command-line arguments (as shown in the \texttt{args} field).

\textbf{1. Using the local Python library:}

\begin{lstlisting}[language=json]
{
  "mcpServers": {
    "paddleocr-ocr": {
      "command": "paddleocr_mcp",
      "args": [
        "--device", "gpu:1"
      ],
      "env": {
        "PADDLEOCR_MCP_PIPELINE": "OCR",
        "PADDLEOCR_MCP_PPOCR_SOURCE": "local"
      }
    }
  }
}
\end{lstlisting}

\textbf{2. Using a PaddlePaddle AI Studio service:}

\begin{lstlisting}[language=json]
{
  "mcpServers": {
    "paddleocr-ocr": {
      "command": "paddleocr_mcp",
      "args": [
        "--timeout", "60"
      ],
      "env": {
        "PADDLEOCR_MCP_PIPELINE": "OCR",
        "PADDLEOCR_MCP_PPOCR_SOURCE": "aistudio",
        "PADDLEOCR_MCP_SERVER_URL": "<server-url>", 
        "PADDLEOCR_MCP_AISTUDIO_ACCESS_TOKEN": "<access-token>"
      }
    }
  }
}
\end{lstlisting}

\textbf{3. Using a self-hosted service:}

\begin{lstlisting}[language=json]
{
  "mcpServers": {
    "paddleocr-ocr": {
      "command": "paddleocr_mcp",
      "args": [],
      "env": {
        "PADDLEOCR_MCP_PIPELINE": "OCR",
        "PADDLEOCR_MCP_PPOCR_SOURCE": "self_hosted",
        "PADDLEOCR_MCP_SERVER_URL": "<server-url>"
      }
    }
  }
}
\end{lstlisting}

\setcounter{figure}{0}
\makeatletter 
\renewcommand{\thefigure}{A\@arabic\c@figure}
\makeatother

\setcounter{table}{0}
\makeatletter 
\renewcommand{\thetable}{A\@arabic\c@table}
\makeatother

\end{document}